\documentclass[conference]{IEEEtran}
\usepackage{times}

\usepackage[numbers]{natbib}
\usepackage{multicol}
\usepackage[bookmarks=true]{hyperref}
\hypersetup{
    colorlinks=true,
    linkcolor=blue,
    filecolor=magenta,      
    urlcolor=cyan
    }
\usepackage{graphics} 
\usepackage{epsfig} 
\usepackage{mathptmx} 
\usepackage{amsmath} 
\usepackage{amssymb}  
\usepackage{siunitx} 
\usepackage{textcomp}
\usepackage{gensymb} 
\usepackage{booktabs}
\usepackage{multirow}
\usepackage{xcolor}
\usepackage{algorithm}
\usepackage{algpseudocode}
\let\labelindent\relax
\usepackage{enumitem}
\usepackage{xspace}
\usepackage{float}
\usepackage{comment}
\usepackage{mathtools}
\usepackage{soul}
\usepackage{natbib}

\usepackage{makecell}
\usepackage{rotating}
\usepackage[percent]{overpic}
\usepackage{contour}
\usepackage{courier}

\usepackage{subcaption} 
\usepackage[font=small]{caption}
\usepackage[percent]{overpic}

\usepackage[11pt]{moresize}

\usepackage{pifont}
\definecolor{MyDarkBlue}{rgb}{0,0.08,1}
\definecolor{airforceblue}{rgb}{0.36, 0.54, 0.66}
\definecolor{MyDarkGreen}{rgb}{0.02,0.6,0.02}
\definecolor{MyDarkRed}{rgb}{0.8,0.02,0.02}
\definecolor{MyDarkOrange}{rgb}{0.40,0.2,0.02}
\definecolor{MyPurple}{RGB}{111,0,255}
\definecolor{MyRed}{rgb}{1.0,0.0,0.0}
\definecolor{MyGold}{rgb}{0.75,0.6,0.12}
\definecolor{MyDarkgray}{rgb}{0.66, 0.66, 0.66}
\definecolor{MyPink}{rgb}{0.9, 0.33, 0.5}
\definecolor{MyCyan}{rgb}{0., 0.4, 0.4}

\definecolor{guidance_green}{RGB}{12,131,27}
\definecolor{theme_orange}{RGB}{255,138,0}
\definecolor{theme_blue}{RGB}{67,99,216}
\definecolor{theme_taro}{RGB}{219,176,234}

\definecolor{pure_green}{RGB}{0,255,0}
\definecolor{pure_red}{RGB}{255,0,0}

\makeatletter
\DeclareRobustCommand\onedot{\futurelet\@let@token\@onedot}
\def\@onedot{\ifx\@let@token.\else.\null\fi\xspace}

\def\eg{\emph{e.g}\onedot} 
\def\ie{\emph{i.e}\onedot}

\def\wrt{w.r.t\onedot} 

\makeatother

\newcommand{\ours}{RoboPanoptes\xspace}

\usepackage[most]{tcolorbox}

\usepackage{xparse}
\usepackage{lipsum}

\def\exampletext{Example} 

\NewDocumentEnvironment{testexample}{ O{} }
{
    \colorlet{colexam}{theme_blue} 
    \newtcolorbox[use counter=testexample]{testexamplebox}{%
        empty,
        title={\exampletext: #1},
        attach boxed title to top left,
        minipage boxed title,
        boxed title style={empty,size=minimal,toprule=0pt,top=2pt,left=2mm,overlay={}},
        coltitle=colexam,fonttitle=\bfseries,
        before=\par\nonskip\noindent,parbox=false,boxsep=0pt,left=2mm,right=0mm,top=2pt,breakable,pad at break=0mm,
        before upper=\csname @totalleftmargin\endcsname0pt, 
        after=\par\nonskip,
        overlay unbroken={\draw[colexam,line width=.5pt] ([xshift=-0pt]title.north west) -- ([xshift=-0pt]frame.south west); },
        overlay first={\draw[colexam,line width=.5pt] ([xshift=-0pt]title.north west) -- ([xshift=-0pt]frame.south west); },
        overlay middle={\draw[colexam,line width=.5pt] ([xshift=-0pt]frame.north west) -- ([xshift=-0pt]frame.south west); },
        overlay last={\draw[colexam,line width=.5pt] ([xshift=-0pt]frame.north west) -- ([xshift=-0pt]frame.south west); },%
    }
    \begin{testexamplebox}}
        {\end{testexamplebox}\endlist}

\AfterEndEnvironment{testexample}{\color{black}}

\DeclareMathAlphabet{\mathcal}{OMS}{cmsy}{m}{n}

\begin{document}
\title{RoboPanoptes: \\ The All-seeing Robot with Whole-body Dexterity}

\author{
Xiaomeng Xu$^{1}$~~~
Dominik Bauer$^{2}$~~~
Shuran Song$^{1,2}$~~~
\\
$^1$Stanford University~~~
$^2$Columbia University~~~
\vspace{0.3cm}
\\
\href{https://robopanoptes.github.io}{https://robopanoptes.github.io}
}
\vspace{-4mm}

\twocolumn[{%
\renewcommand\twocolumn[1][]{#1}%
\maketitle
\includegraphics[width=\linewidth]{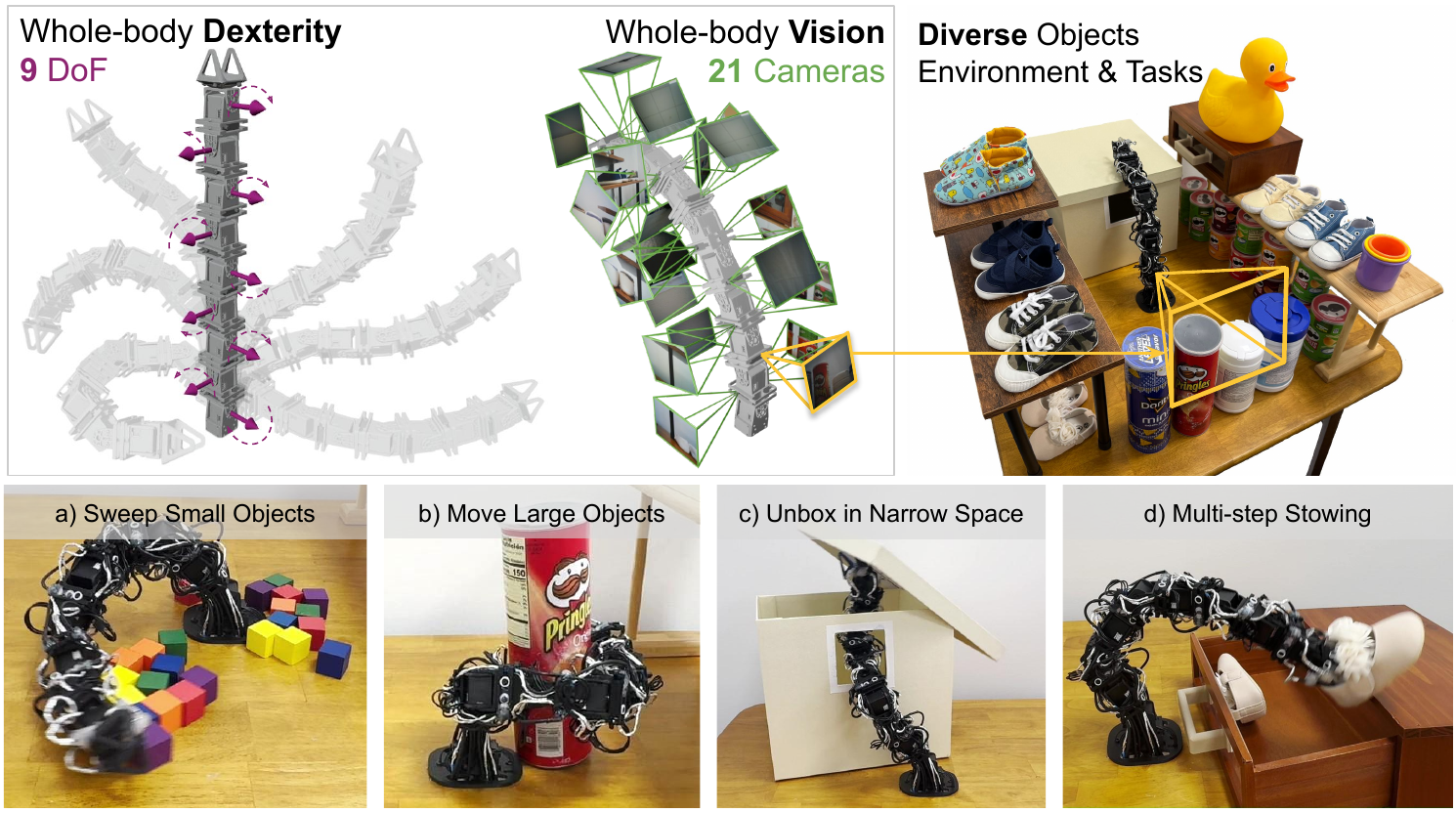}
\captionof{figure}{
\textbf{\ours}, a robot that utilizes all of its body parts to \textbf{sense} and \textbf{interact} with its environment. Whole-body vision (via 21 cameras distributed over the robot's body) enables whole-body dexterity, with the robot utilizing its entire surface for manipulation. 
This design enables new robot capabilities such as a) simultaneously sweeping multiple small objects, b) moving large objects using whole-body contact, c) unboxing in constrained, narrow spaces, and d) executing precise multi-step stowing in cluttered environments.
\vspace{1.5em}
\label{fig:task}}
}]

\begin{abstract}
We present RoboPanoptes\footnote{%
\textit{Argos Panoptes} is a many-eyed giant in Greek mythology, with ``panoptes'' meaning ``all-seeing''. \textit{RoboPanoptes} is the many-eyed, all-seeing robot.%
}, a capable yet practical robot system that achieves whole-body dexterity through whole-body vision. 
Its whole-body dexterity allows the robot to utilize its entire body surface for manipulation, such as leveraging multiple contact points or navigating constrained spaces.
Meanwhile, whole-body vision uses a camera system distributed over the robot's surface to provide comprehensive, multi-perspective visual feedback of its own and the environment's state. 
At its core, RoboPanoptes uses a whole-body visuomotor policy that learns complex manipulation skills directly from human demonstrations, efficiently aggregating information from the distributed cameras while maintaining resilience to sensor failures. 
Together, these design aspects unlock new capabilities and tasks, allowing RoboPanoptes to unbox in narrow spaces, sweep multiple or oversized objects, and succeed in multi-step stowing in cluttered environments, outperforming baselines in adaptability and efficiency. Results are best viewed on \href{https://robopanoptes.github.io}{https://robopanoptes.github.io}.


\end{abstract}

\IEEEpeerreviewmaketitle


\section{Introduction}
The vast majority of today's robot manipulation systems focus on controlling the robot's ``end-effector'' (\eg, a hand or gripper) using a ``centralized camera'' (\eg, an environment- or wrist-mounted camera). Despite being the common practice, both designs unnecessarily limit a robot's capability.  In this paper, we challenge these conventional designs by introducing \textbf{\ours}, a novel robot system that achieves {\textbf{whole-body dexterity}} through {\textbf{whole-body vision}}.

By whole-body dexterity, we mean the manipulation skills that utilize the robot's entire body surface to interact with the environment. For example, the robot uses all its links to sweep many objects at once (Fig.~\ref{fig:task}~a), instead of using only the end-effector to push objects one by one. To demonstrate the benefit of whole-body dexterity, we design a high degree-of-freedom robot, where the redundancy of the degrees of freedom (DoF) allows high motion flexibility and adaptability.

However, achieving whole-body dexterity requires the robot to \textit{coordinate} all its movable body parts, \textit{understand} their spatial relationships with respect to the environment, and \textit{plan} their interactions therewith. Attaining the required level of perception is challenging with a single centralized camera, regardless of its placement. To address this challenge, we introduce the concept of {whole-body vision} -- an integrated camera system that is distributed on the robot’s surface, providing comprehensive visual feedback for whole-body dexterity.

Together, these design aspects unlock new capabilities for robotic manipulation systems, including:  

\begin{itemize}[leftmargin=3.5mm]
\item  \textbf{Multi-contact manipulation} using all body parts, such as simultaneously sweeping multiple objects or moving an object larger than the robot itself (Fig.~\ref{fig:task}~a,~b).

\item  \textbf{Manipulation in constrained spaces}, such as unboxing a container from its side opening or organizing objects in a cluttered environment (Fig.~\ref{fig:task}~c,~d).

\item  \textbf{Omni-directional manipulation}, allowing the robot to perceive and manipulate its environment from any angle without reorienting its body, thereby increasing its efficiency in all tasks by maximizing the share of task-related motion.
\end{itemize}

However, the increased dexterity and perception complexity create challenges for controlling such a system, and conversely, new opportunities in the system's design:

\textbf{Control complexity:}
Programming complex behaviors for such a high degree-of-freedom system is demanding. Previous approaches often rely on manually designed state representations and carefully tuned motion patterns (\eg, sinusoid functions~\cite{liljeback2012review, wright2007design}) for every task, which are not transferable for new task requirements. We drastically simplify this process by introducing an intuitive demonstration interface and an efficient policy learning algorithm, allowing even novice users to quickly and easily teach the robot new manipulation skills. 

\textbf{Learning efficiency:}
For whole-body vision, numerous cameras are distributed on the robot's surface and move in tandem with the robot. Consequently, the policy must efficiently process this complex and high-dimensional input space to infer the appropriate actions. We address this challenge through a novel ``whole-body visuomotor policy'' architecture based on diffusion transformers~\cite{chi2023diffusion}. Cross-attention captures semantic correlations between multi-view images and robot actions, while a view-dependent positional encoding preserves crucial spatial information. As we demonstrate for diverse tasks, this architecture enables efficient learning of complex behaviors from limited data.

\textbf{System robustness:} 
Multi-sensor systems often become increasingly vulnerable to failure as the number of sensors grows, where a single camera malfunction may generate out-of-distribution observations that compromise the overall system performance. To address this challenge, we develop a ``Blink Training'' strategy that randomly masks camera inputs during the training process. By simulating these unpredictable sensor disruptions -- analogous to random ``blinking'' of the robot's cameras -- we are able to train policies that remain reliable even under partial sensor failure or latency.

In summary, our primary contribution is the \textbf{\ours} system, demonstrating novel whole-body dexterity capabilities through whole-body vision. Specifically, it is enabled by: 

\begin{itemize}[leftmargin=4mm]
\item A \textbf{modular hardware design} that integrates distributed vision and actuation in a scalable and reconfigurable system.  

\item An \textbf{intuitive demonstration interface} that streamlines the acquisition of complex whole-body manipulation behaviors, dramatically reducing the barrier to teaching new skills.

\item A \textbf{whole-body visuomotor policy} that efficiently processes whole-body visual input through cross-attention transformers and view-dependent positional encoding, while improving resilience to sensor failures through blink training.
\end{itemize}

Our hardware design, training data, teleoperation and policy code are publicly available on \href{https://github.com/real-stanford/RoboPanoptes}{https://github.com/real-stanford/RoboPanoptes}. 

\section{Related work}

By discussing prior work on designing high-DoF robots, on leveraging them for whole-body manipulation and the closely related challenge of whole-body sensing, we illustrate the novel capabilities of \ours with regards to achieving whole-body \textit{dexterity} by exploiting whole-body \textit{vision}.

\subsection{High Degree-of-Freedom Robot Systems}
High-DoF robotic systems are characterized by their ability to achieve complex configurations and motions through a large number of controllable joints. This hyper-redundancy enables them to emulate their biological role models -- such as snakes, vines~\cite{coad2019vine, blumenschein2020design}, and elephant trunks~\cite{zhang2023preprogrammable} -- to perform tasks in confined and complex environments.
Snake robots~\cite{liljeback2012review, wright2007design, qin2022adaptive, seeja2022survey,liu2021review, wang2023mechanical, jiang2024hierarchical,jiang2024snake} are especially prominent examples of such high-DoF systems. Leveraging hyper-redundancy, they are able to achieve undulating, side-winding, and tumbling movement strategies. 
While snake robots demonstrate these impressive locomotion capabilities, their use in manipulation remains limited, often restricted to simple grasping~\cite{elsayed2021mobile, liu2023inchworm, salagame2024loco, cai2025modular}, hard-coded control strategies~\cite{wright2007design, elsayed2021mobile}, or optimization-based path planning with manual primitive designs~\cite{salagame2024loco}.
In contrast, our work extends the potential of high-DoF systems by incorporating whole-body vision, intuitive data collection interfaces, and visuomotor policies, enabling robust, dexterous manipulation across diverse and complex tasks.



\subsection{Whole-body Manipulation}
\citet{stasse:hal-03045448} define whole-body manipulation as requiring high redundancy, a floating base, and multiple contact points. 
Previous whole-body manipulation systems typically consist of a quadruped with an arm mounted on top~\cite{liu2021garbage, jeon2023learning, fu2023deep, ha2024umi}, or a bimanual system with legs or a mobile base~\cite{ren2022integrated, fu2024mobile, sferrazza2024humanoidbench, xiong2024adaptive, shaw2024bimanual, uppal2024spin}. However, while existing systems optimize body configurations to achieve desired end-effector trajectories, they still fundamentally \textbf{restrict environmental contact to the end-effector alone}. In contrast, whole-body dexterity actively leverages all available body surfaces for environmental contact and manipulation (with or without a movable base).
Moreover, the sensing capabilities of existing whole-body manipulation systems are limited. By commonly using a single \mbox{environment-}, wrist- or head-mounted camera, occlusions are unavoidable and hinder their ability to fully observe the state of cluttered environments and the robot's state relative to it.

\subsection{Whole-body Sensing}
Prior work on whole-body sensing has explored range, tactile, and force sensing methods to enhance robot perception and interaction, addressing challenges in collision avoidance, contact detection, and compliant motion.
Range sensing enables collision avoidance and spatial awareness in manipulation systems. \citet{tanaka2015range} implement a semi-autonomous collision avoidance system for snake robots using range sensors distributed along the body, enabling it to navigate complex terrains. 
\citet{qi2022safe} further develop a laser-ranging sensor ring design for human-robot interaction.
\citet{kim2024armor} employ Time-of-Flight sensors distributed on a humanoid robot for collision-free motion planning.
Whole-body force sensing, as implemented by \citet{kollmitz2018whole}, leverages force-torque sensors embedded within mobile robots to achieve compliant motion.
Tactile sensors allow robots to interact delicately with the environment by detecting pressure, texture, and force~\cite{dahiya2009tactile,liu2023enhancing,dean2019whole,jiang2024hierarchical}. 
Most noticeably, the Punyo robot \cite{goncalves2022punyo} uses various contact and force sensors throughout the robot's chest, arm, and hand to enable different resolutions of contact signals across the body, which is crucial for contact-rich whole-body manipulation tasks.

While effective in detecting \textit{nearby} obstacles and contacts, the understanding of the \textit{semantics} of the environment remains limited using range, tactile, and force sensing
. In contrast, \textbf{vision} provides rich information about objects and environments beyond contact or geometry. 
Most related to our work, \citet{yamaguchi2017optical} propose 
the idea of whole-body vision with an optical skin system; combining embedded cameras with transparent, flexible skin. 
However, their implementation is constrained by high data transmission requirements and integration complexity such as bulky Raspberry Pi cameras and processors. 
In contrast, we fully implemented an effective and robust whole-body vision setup capable of autonomously performing dexterous manipulation skills enabled by our modular hardware design (\S~\ref{sec:design}), intuitive data collection interface (\S~\ref{sec:teleop}), whole-body visuomotor policy (\S~\ref{sec:policy}), as well as several practical considerations during system implementation (\S~\ref{sec:pratical}). Through our experiments, we demonstrate that the whole-body vision data is sufficient for imitation learning of complex manipulation tasks (\S~\ref{sec:experiment}).


\section{Design Objectives}

We design \ours to be \textbf{capable yet practical}; able to perform a wide range of whole-body dexterity tasks, while remaining easy to build, extend, and deploy.

\vspace{2mm}
\noindent Specifically, \ours' \textbf{capabilities} are characterized by:

\begin{itemize}[leftmargin=3.5mm]
    \item \textit{Motion flexibility}, enabling the system to leverage all body parts for object manipulation and collision avoidance.
    \item \textit{Visual coverage}, minimizing occlusions caused by the environment or the robot's body itself.
    \item \textit{Rapid adaptability} to new tasks and applications without explicit reprogramming, instead directly leveraging human demonstration.   
\end{itemize}

\vspace{2mm}
\noindent The \textbf{practicality} of \ours is reflected in its:
\begin{itemize}[leftmargin=3.5mm]
 
 \item \textit{Robustness} to sporadic sensor delays and failures; an issue that compounds as the number of cameras increases. 
 \item \textit{Modular design}, allowing easy customization with varying degrees of freedom (DoFs) and camera configurations.
 \item \textit{Reproducibility}, as the system is low-cost and built with off-the-shelf and 3D printed components, enabling researchers to easily replicate it.
\end{itemize}

We achieve these objectives through a series of careful hardware design decisions \S~\ref{sec:design}, an intuitive data collection interface \S~\ref{sec:teleop}, and robust visuomotor policy learning \S~\ref{sec:policy}. In~\S~\ref{sec:pratical}, we also discuss important practical considerations for designing whole-body vision systems.


\begin{figure}[h]
    \centering
    \includegraphics[width=\linewidth]{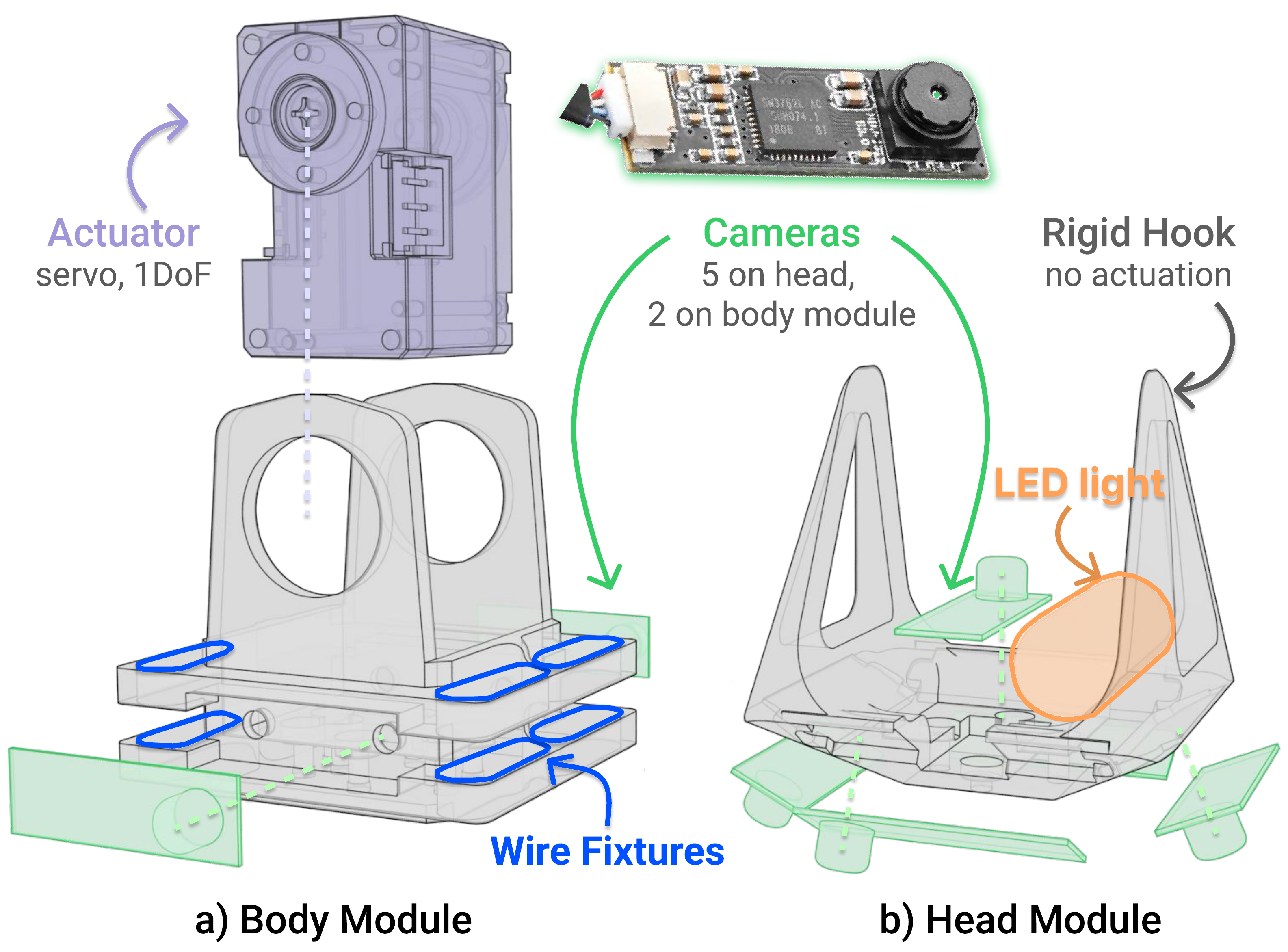}
    \caption{
    \textbf{Modular Hardware Design} including a) a body module consisting of an actuator, two cameras, and wire fixtures, as well as b) a head module with five cameras and an LED light. 
    }
    \vspace{-4mm}
    \label{fig:design}
\end{figure}

\section{Modular Hardware Design}
\label{sec:design}
\ours' hardware consists of nine modular body units and one head unit. The design objectives achieved thereby are extendability to higher DoFs, comprehensive visual coverage, and ease of (hardware) reproducibility.

\vspace{2mm}\noindent\textbf{Vision-Actuation Body Module.}
Each body module (shown in Fig.~\ref{fig:design}~b) contains an actuator, two cameras, and fixtures for wire management.
The actuator is a standard \texttt{DYNAMIXEL XC330-M288-T} servo motor with one rotational degree of freedom. This motor is small ($20\times34\times26$ \SI{}{mm}) and lightweight (\SI{23}{g}), with a 288.4:1 gear ratio. These motors are daisy-chained together; a single wire powers the robot. We control them with current-based position control communicated through the TTL protocol. The body modules are chained together, with neighboring links' DoF perpendicular to each other, as shown in Fig.~\ref{fig:task}. Each motor can freely rotate up to \SI{\pm90}{\degree}, allowing flexible and complex motions of the chain of body modules.
 
To maximize the visual coverage while minimizing the number of cameras, we mount two cameras on each link -- one on the front and one on the back -- with cameras pointing outwards and perpendicular to the link's DoF. Using eight body modules, the whole robot is thus equipped with 16 cameras -- four cameras on each of its four sides --, enabling omnidirectional (\textit{``all-seeing''}) vision. We design the camera slots on the links to hold the cameras mechanically stable to avoid relative movement \wrt the link, and caging the cameras inside the body to avoid accidental damage. We use the \texttt{Adafruit Ultra Tiny USB Camera} with a \texttt{GC0307} image sensor, characterized by an extremely small dimension ($8\times25\times4.5$ \SI{}{mm}) that easily fits on the small links. It offers a sufficient resolution of $640\times480$ and a \SI{50}{\degree} field of view. However, each camera requires an adapter cable that converts the camera board's JST connector to a USB-A port, and the cameras cannot be daisy-chained. If not managed properly, the resulting large number of cables would constrain the robot's motion and occlude camera views. To address this, we design wire fixtures that are incorporated on each link. These slots allow wires to pass through on the remaining two sides of the module that are not holding cameras, ensuring that the wires will not occlude the cameras.

\vspace{2mm}\noindent\textbf{Head Module.} The final body module is connected to a head module that acts as an end-effector to unlock more manipulation skills and provides additional perception capabilities. 
We design this module in the shape of a parallel hook that enables the picking and pulling of small extrusions, \eg, hooking and pulling the drawer's handle in Fig.~\ref{fig:task}~b.
Even though the robot's body modules observe all four sides around it, some blind angles (facing top and down) remain, which we address through the head module's camera mounting strategy. Four downwards-facing cameras are placed on the lower part of the unit, providing additional top-down views. Another camera is placed in the center of the head and pointing upwards.
Moreover, a battery-powered LED light is placed next to the upward camera, illuminating the area that the head module navigates towards and that its hooks might interact with. This allows the robot to visually perceive even narrow dark spaces, \eg, inside the box in Fig.~\ref{fig:task}~a.
Alternatively, this head design may be flexibly swapped with other types of end-effectors for specific tasks.

\begin{figure}[h]
    \centering
    \includegraphics[width=\linewidth]{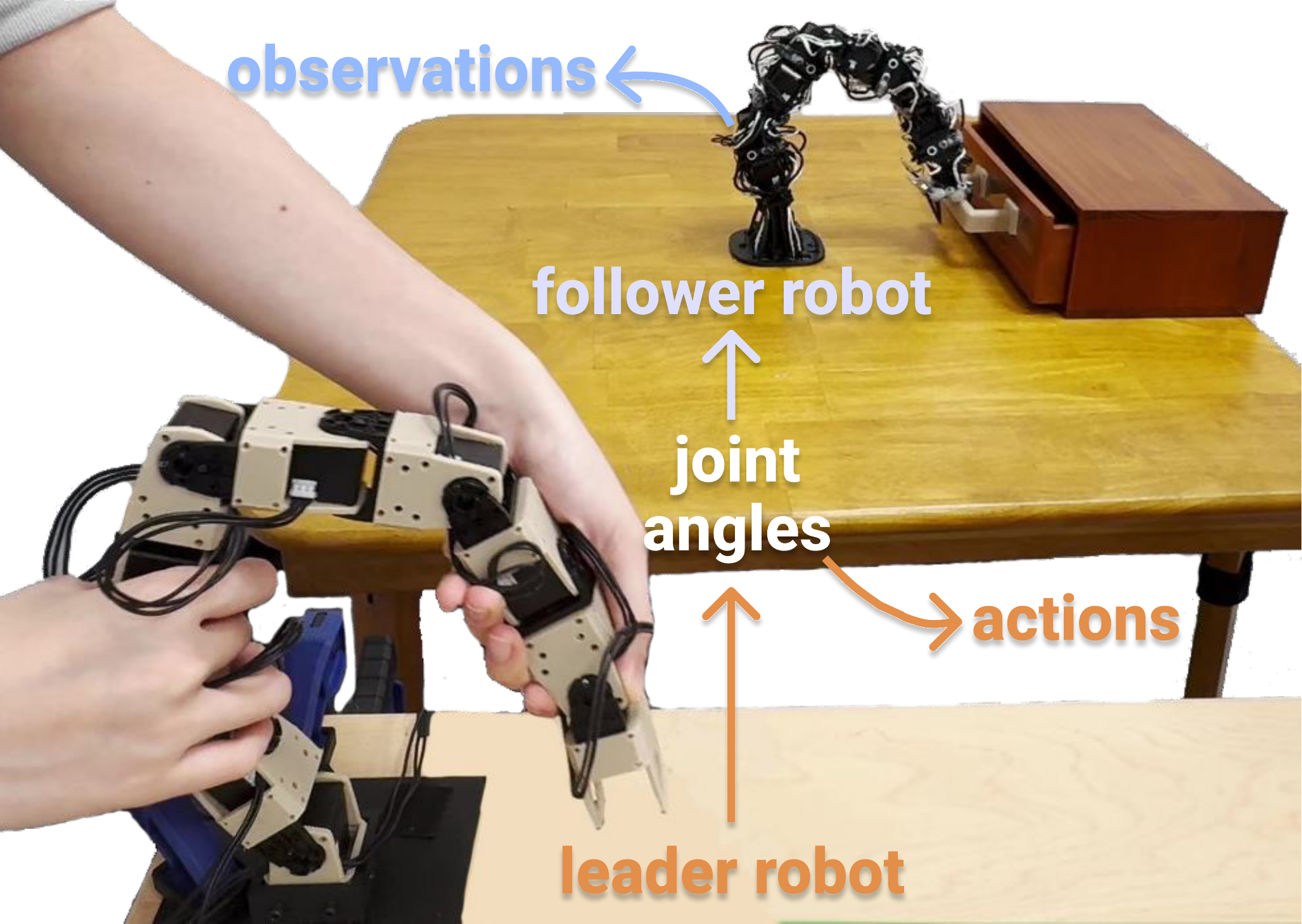}
    \caption{\textbf{Data Collection Interface.} The operator uses both hands to control the leader robot, whose joint angles are sent to the follower robot in real-time as position targets. The joint angles of the leader robot are recorded as target actions, while the images and joint angles of the follower robot are recorded as observations.
    }
    \label{fig:teleop}
    \vspace{-4mm}
\end{figure}
\section{Data Collection Interface}
\label{sec:teleop}

Directly programming such a high-DoF robot to execute complex (and natural) behaviors is difficult. Inverse kinematics and standard joystick controls, commonly used in manipulation systems, can only control end-effector motions and are hard to solve for high-DoF systems. For locomotion of similarly structured high-DoF robots, previous approaches use carefully coded motion patterns (\eg, sinusoid functions~\cite{liljeback2012review, wright2007design, cai2025modular}), which do not transfer to complex manipulation tasks.
While RL exploration offers an alternative, it can only be safely and scalably implemented in simulation, introducing significant sim2real observation gaps. 

Instead, to enable intuitive control and \textbf{adaptability} to new skills, we design a leader-follower teleoperation system~\cite{wu2024gello} as shown in Fig.~\ref{fig:teleop}. The leader and the follower robot have the exact same structure. During teleoperation, torque is disabled for the leader robot while being enabled for the follower. To demonstrate a task, a human operator uses both hands to move the leader robot. The leader's joint positions are sent to the follower in real time, allowing it to mirror the leader using PID position control at a control rate of 30 Hz. During the demonstration, images from all cameras and robot joint positions are recorded at 10 Hz.
%

\begin{figure*}[t]
    \centering
    \includegraphics[width=1.0\linewidth]{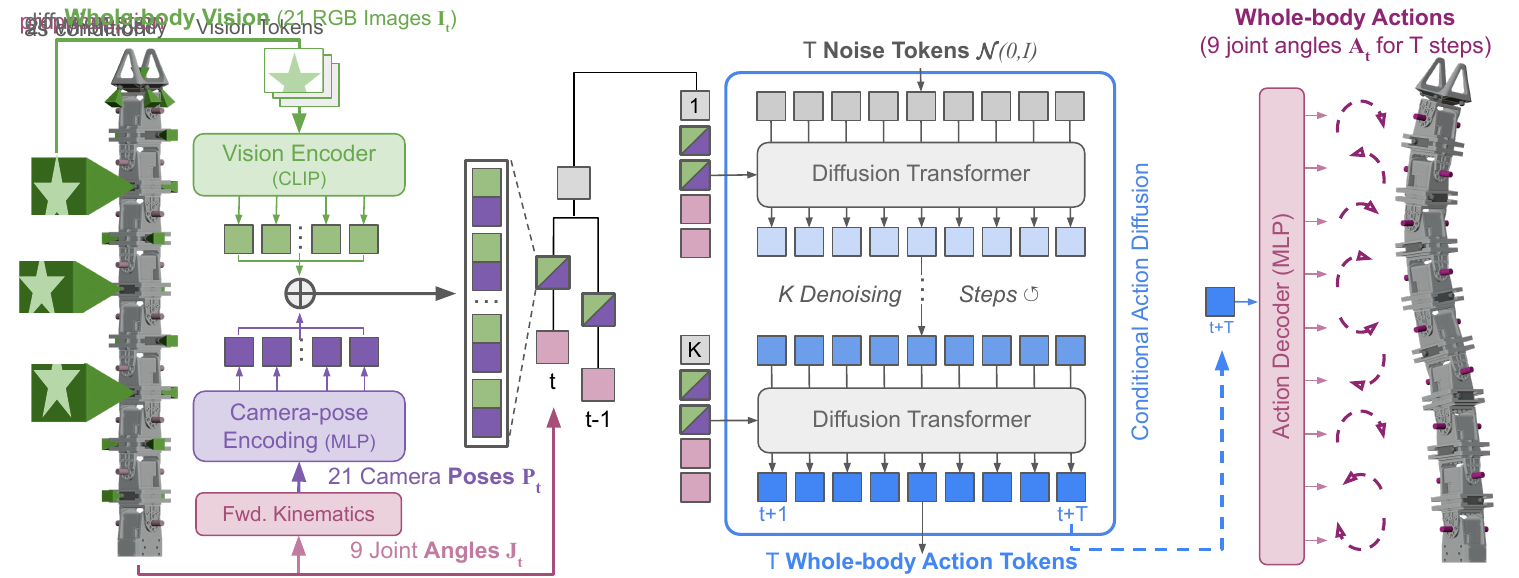}
    
    \caption{\textbf{Whole-body Visuomotor Policy} leverages whole-body vision for whole-body dexterity.  Left: The current robot and environment state is observed via RoboPanoptes' 21 cameras and its 9 joints angles, converted to 21 camera poses using forward kinematics. Each image (green) is represented by the class token of a vision foundation model. Each camera pose (purple) is embedded using our view-dependent positional encoding. The concatenation of each camera's image and pose tokens yields a whole-body vision token; 21 in total. Middle: Our whole-body visuomotor policy consumes these vision tokens, proprioception, and denoising-step tokens as condition via cross attention. We diffuse $T$ whole-body dexterity tokens (blue), each corresponding to an action time step. Right: Per time step, we project the predicted dexterity token to the 9 joint angles to be achieved by the dexterity action.}
    \label{fig:policy}
    \vspace{-4mm}
\end{figure*}
\section{Whole-body Visuomotor Policy}
\label{sec:policy}


Using the collected demonstrations, we can train a whole-body visuomotor policy that infers whole-body actions (\ie, nine joint angle sequences) given whole-body vision (\ie, images from 21 cameras). Compared to a common manipulation system, \ours needs to handle significantly more complex observation spaces due to the following factors: 

\begin{itemize}[leftmargin=4mm]
   \item \textbf{Many cameras:} The policy needs to efficiently extract task-relevant semantic information from the raw pixels captured by a large number of cameras. 
    
    \item \textbf{Non-stationary cameras:} As the robot moves, the cameras are in constant motion. This requires the policy to understand the {dynamic spatial relations} between the cameras, the robot and its environment. 
    
    \item \textbf{Unreliable cameras:} A system of many cameras is prone to unpredictable failures and delays, requiring the policy to be robust to such disturbances. 
\end{itemize}

As shown in Fig.~\ref{fig:policy}, we base our policy design on Diffusion Policy~\cite{chi2023diffusion}, which infers the robot's joint-space actions through denoising diffusion, conditional on previous observations. Concretely, at each time step $t$, the policy takes the previous $T_{o}$ observations~$\mathbf{O}_{t}$ as conditional input. It predicts $T_{p}$ actions~$\mathbf{A}_t$, of which $T_a\leq T_p$ are executed on the robot. In our implementation, we set the observation horizon $T_o=2$, the action prediction horizon $T_p=16$, and the action execution horizon $T_a=8$. The observations $\mathbf{O}_{t}=\{\mathbf{I}_t,\mathbf{P}_t,\mathbf{J}_t\}$, taken from the follower robot during training, consist of the previous RGB observations $\mathbf{I}_t=\{i_{t-T_o+1},...,i_t\}\in\mathbb{N}_0^{T_o\times21\times640\times480\times3}$ from 21 cameras, their poses $\mathbf{P}_t=\{p_{t-T_o+1},...,p_t\}\in\mathbb{R}^{T_o\times21\times(3+6)}$, and the corresponding joint angles $\mathbf{J}_t=\{j_{t-T_o+1},...,j_t\}\in\mathbb{R}^{T_o\times9}$. The actions, taken from the leader robot during training, are the following joint angles $\mathbf{A}_t=\{a_{t+1},...,a_{t+T_p}\}\in\mathbb{R}^{T_p\times9}$. 

In the following, we elaborate on the policy design details and outline how these key design decisions address the aforementioned challenges.

\vspace{2mm}\noindent\textbf{Coordinating moving cameras with view-dependent positional encoding.}
The robot's motions continuously change the camera poses and, thereby, the region of the environment each camera is observing. The observation space of this \textit{dynamic} multi-view camera system is thus significantly larger than that of, \eg, a static top-down camera. To coordinate the moving cameras and make policy learning more efficient, we employ a view-dependent positional encoding strategy.
Specifically, we compute the 6D camera poses in the base frame using the forward kinematics based on the current joint angles. Each camera pose is represented by a $3$-dimensional vector for position and a $6$-dimensional vector for orientation, using the first two columns of rotation matrix~\cite{zhou2019continuity}. The position and orientation vectors are each projected to a $192$-dimensional vector using fully-connected linear layers. 
Our experiments demonstrate that this additional positional encoding enables the policy to have better data efficiency and 3D awareness.

\vspace{2mm}\noindent\textbf{Learning semantic correspondences using a multi-view cross-attention transformer.}
To efficiently extract task-relevant information from many cameras, we employ a pretrained vision encoder and a multi-view cross-attention transformer to learn the task-relevant semantic correspondence between different camera observations.
To this end, we exploit pretrained vision foundation models such as CLIP~\cite{radford2021learning} or DINO~\cite{caron2021emerging} that enable advanced semantic understanding~\cite{xu2023jacobinerf} and visually-complex robot manipulation tasks~\cite{chi2024universal, xu2024flow}.
Specifically, we first resize the images to $224\times224$ resolution and apply color jitter augmentation. We batch the 21 images and feed them into the (frozen) ViT-B/16~\cite{dosovitskiy2020image} encoder of a CLIP-pretrained model to predict a $768$-dimensional class-token feature for each image. These per-image features are projected to $384$-dimensional vectors and concatenated with the corresponding camera's position and orientation embeddings, resulting in a $768$-dimensional token per camera. The concatenation of these whole-body vision tokens with the projected joint angles yields a $(21+9)\times768$-dimensional representation of the robot and environment state at time $t$. 

We adopt a time-series diffusion transformer architecture~\cite{chi2023diffusion}.
Noisy actions $\mathbf{A}^{k}_t$ (where $k$ is the denoising step) are passed in as input tokens for the transformer decoder blocks, while the observations $\mathbf{O}_t$ are passed into the multi-head cross-attention layers of the transformer decoder stack as conditional input.
The cross-attention mechanism~\cite{vaswani2017attention, chi2023diffusion} efficiently leverages the rich information captured by all cameras, learning the correspondences between multi-modal observations (multi-view images and robot proprioception) and robot actions.
%

\vspace{2mm}\noindent\textbf{Robustness to unreliable cameras with blink training.}
Low-cost cameras often have unreliable connections, leading to dropouts and variable latencies. On average, the dropout rate (\ie, camera discount) observed for the used cameras is 4.4\% and latency ranges from \SI{15}{ms} to \SI{100}{ms}.
To make the system robust to such camera failure at test time, we employ a ``blink training'' strategy that randomly drops out camera inputs during training.
Concretely, we simulate a 5\% failure rate for each camera, \textit{independently} masking out entire images at each time step. This means that there is a 65.9\% probability that \textit{at least one} camera is being dropped out at each time step.
Consistent with observations in previous work~\cite{skandsimple}, this simple strategy significantly improves the robustness of the policy, enabling it to succeed even when some cameras are completely disabled during test time.

\section{Practical Considerations}
\label{sec:pratical}

This section highlights critical implementation details for developing an effective \ours system. Although we do not consider these aspects to be novel technical contributions, they are crucial to ensuring robust and efficient system performance. As such, our insights provided here may serve as guidelines for the future development of whole-body dexterity and vision systems.

\vspace{1mm}\noindent\textbf{Camera Selection.}
When selecting a camera, we need to consider its size, field of view (FOV), and connectivity. We choose the \texttt{Adafruit Ultra Tiny USB Camera} due to its versatility and ease of integration. It offers a compact design ($8\times25\times4.5$ \SI{}{mm}) that fits well within the spatial constraints of the robot while providing a reasonable FOV ($50\degree$) for the tasks. Moreover, these USB cameras support standard UVC (USB Video Class) interfaces, making them plug-and-play compatible with various devices. This simplifies the deployment and ensures ease of integration.

We also explored other options, such as wireless cameras and Raspberry Pi cameras, but encountered impractical limitations. Wireless cameras introduce significant latency, limiting the reactivity of the overall system during deployment. The \texttt{Raspberry Pi Camera Module} requires a Raspberry Pi single-board computer as the interface, and we find the used ribbon cable connection to be sufficiently flexible but unstable, leading to potential reliability issues. In contrast, USB cameras provide a reliable and standardized interface and, through UVC, are compatible across a wide range of devices without the need for specialized (interface) hardware.

\vspace{1mm}\noindent\textbf{Wire Management.}
Managing the USB and power wires for all 21 cameras and 9 motors turns out to be a critical engineering challenge. It directly impacts the overall system performance as poor wire management causes a range of issues, including obstructed camera views and restricted robot motion. Tangled wires also cause physical strain and potentially lead to an unreliable, or even damaged, system.
To prevent wires from obstructing camera views, we design custom fixtures on the robot's links, as described in \S~\ref{sec:design}, that secure the wires in place.
However, fixing the wires to the links introduces a new issue: the stiffness of the wire could reduce the range of the robot's movements. Note that the USB camera's wiring, which consists of four internal wires (power, ground, data$+$, and data$-$), requires proper insulation to prevent common-mode interference and ensure stable data transmission. However, we find that we can still increase the wires' flexibility by removing the bulky rubber jacket and retaining only the thin aluminum foil below for insulation.


\vspace{1mm}\noindent\textbf{Data Communication Bandwidth.}
A common desktop PC often has limited bandwidth, making it challenging to stream data from multiple high-bandwidth devices, such as cameras. For our choice of camera, the main constraint is the number of USB devices and corresponding video streams a PC can support; usually only four or five simultaneously.
To address this bottleneck, 
we use four PCIe USB extension boards, each providing 20 Gbps of additional bandwidth and allowing the system to reliably stream data from 23 cameras simultaneously.

\vspace{1mm}\noindent\textbf{Multiprocessing.} To efficiently manage data acquisition and processing from multiple cameras, we use multiprocessing to parallelize tasks and minimize interference between camera streams. Each camera is acquired in its own process at 30 FPS. During demonstration and inference, a shared memory space (with the size of $\mathbb{N}_0^{21\times640\times480\times3}$) across all processes is maintained to store the latest image observations, updated every time a new frame is received. This approach ensures seamless data flow and reduces processing delays, allowing the system to handle large amounts of visual data concurrently.



\begin{figure}[h]
    \centering
    \includegraphics[width=\linewidth]{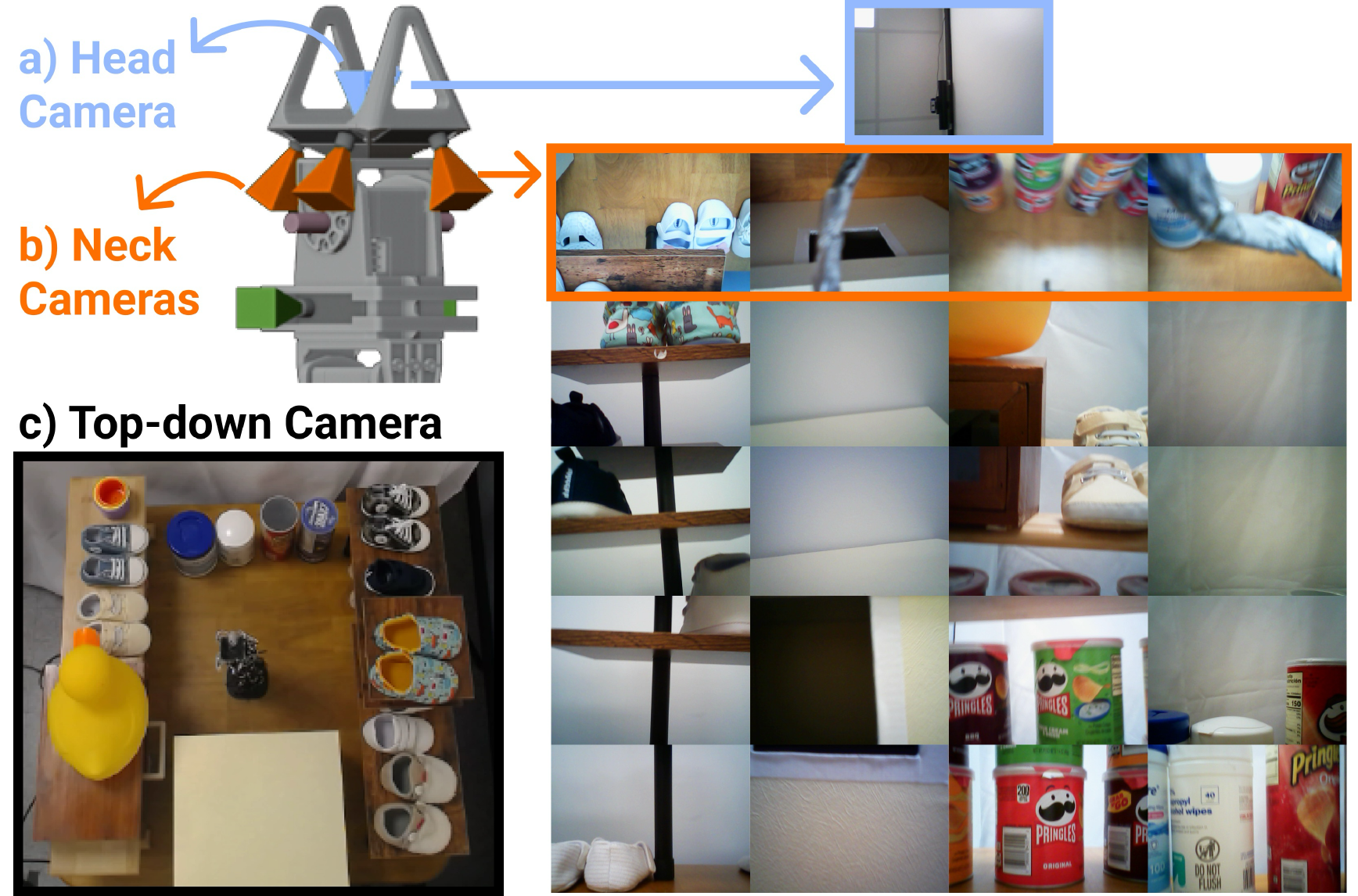}
    \caption{
    \textbf{Alternative Observation Spaces.} We compare \ours with several baselines, including using a) only the head camera, b) the four neck cameras, and c) a top-down camera. Variants using all of \ours' cameras but without view-dependent positional encoding or without blink training serve as ablations of our design.}
    \label{fig:baseline}
    \vspace{-4mm}
\end{figure}

\begin{figure*}[t]
    \centering
    \includegraphics[width=\linewidth]{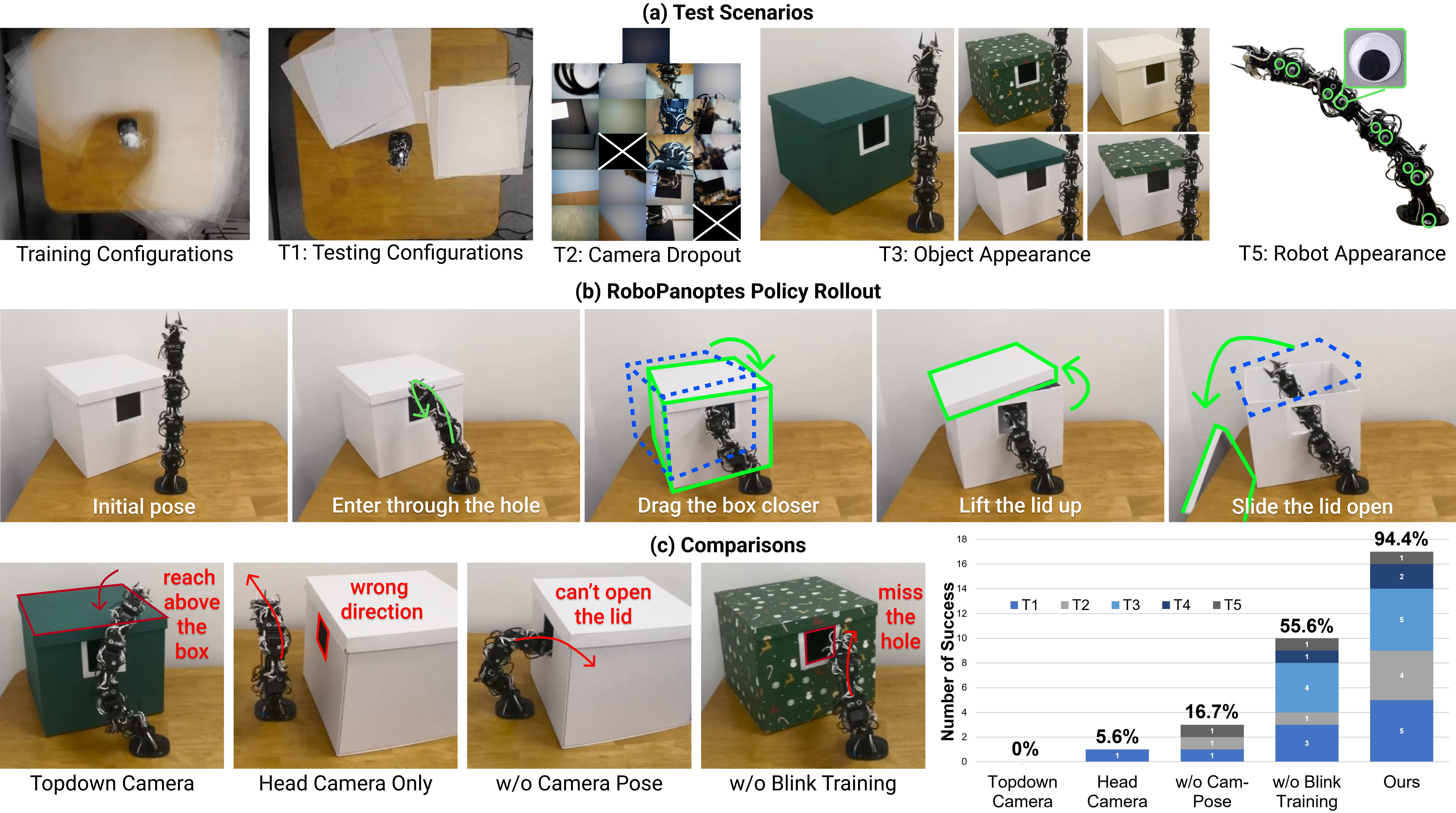}
    \caption{
    \textbf{Unboxing Task.} a) Different test scenarios. The first two images show all initial configurations overlayed. b) Policy rollout of \ours, demonstrating precise manipulation in narrow spaces (inside the box) and utilizing body contacts to drag the box closer. c) Typical failure cases of the baselines. The \texttt{Top-down Camera} policy struggles to determine the correct reaching height. The \texttt{Head Camera} policy moves in the wrong direction. The \texttt{w/o Camera Pose Encoding} policy struggles to open the lid, and at locating and entering the hole. The \texttt{w/o Blink Training} policy underperforms in scenarios with camera dropout (T2). }
    \label{fig:eval_unbox}
    \vspace{-4mm}
\end{figure*}

\section{Experiments}
\label{sec:experiment}
In the following, we study \ours' ability to perform a wide range of real-world manipulation tasks that require whole-body dexterity. To ensure fair comparisons, all methods use the same set of demonstrations for training and identical initial configurations during testing.

\begin{figure*}[t]
    \centering
    \includegraphics[width=\linewidth]{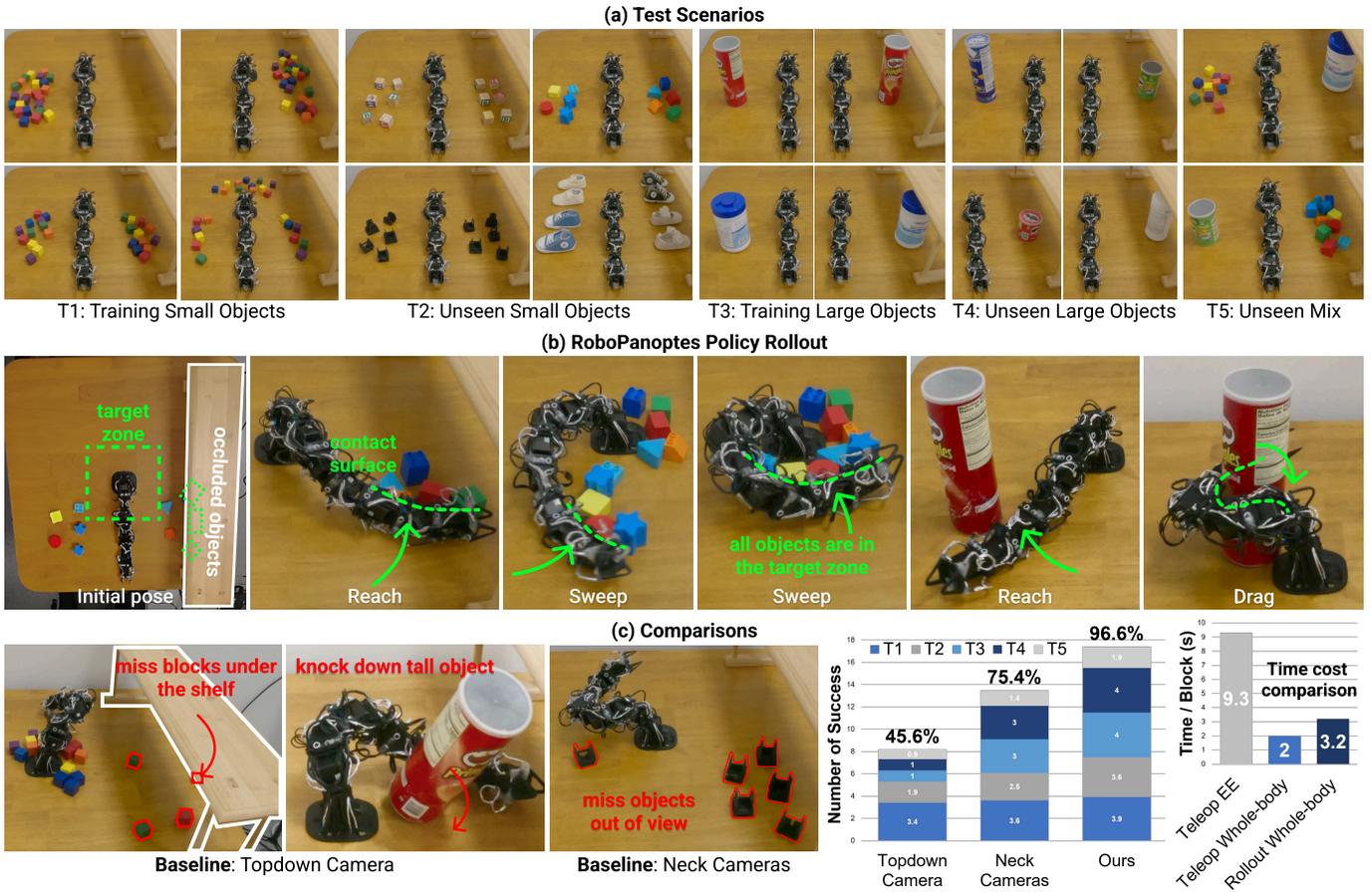}
    \caption{
    \textbf{Sweeping Task.} a) Different test scenarios. b) \ours policy rollouts, highlighting the capability of leveraging multiple whole-body contacts with objects. c) Typical failure cases of baselines. The \texttt{Top-down Camera} policy fails to detect objects under the shelf and often knocks down tall objects. The \texttt{Neck Cameras} policy struggles with objects located behind the robot due to self-occlusion.}
    \label{fig:eval_sweep}
    \vspace{-4mm}
\end{figure*}

\subsection{Unboxing Task}
\vspace{2mm}\noindent\textbf{Task:}
The robot needs to open the lid of a box that is placed in a random location around the robot (Fig.~\ref{fig:eval_unbox}). To do so, the robot first needs to locate the small hole on the side of the box, enter the box through the hole, drag the box closer (if it is out of reach), extend its body inside the box, lift the lid, and finally slide the lid aside to fully open the box. 

The task success is measured by whether the box is opened in the end. No partial credit is given for this task.

\vspace{2mm}\noindent\textbf{Challenges:}
\ul{Constrained space:} The robot must enter the box through a narrow hole and operate within the confined space inside the box. 
\ul{Visual occlusions:} The hole's location is often occluded (with respect to a camera at the scene's center) due to the random box placement.
\ul{Dexterity:} Actions like lifting and sliding the lid require fine-grained control over multiple DoFs involving multiple links.
\ul{Multiple contacts:} The robot must drag the box closer, leveraging multiple contacts with its body links and stabilizing the box during lid manipulation.

\vspace{2mm}\noindent\textbf{Test scenarios:}
We conduct 18 rollouts for each policy. To ensure a fair comparison, we use the same set of initial robot and object configurations for all different methods. We achieve the same object configuration by adjusting it to an overlayed top-down view of the reference configuration. As shown in Fig.~\ref{fig:eval_unbox}~(a), the test configurations can be grouped into five categories:

\begin{itemize}[leftmargin=4mm]
    \item T1: Variation of the initial object pose (5 rollouts).
    \item T2: Random camera dropout during execution, simulating for each camera a 5\% failure rate at each time step (5 rollouts); \ie, a 65.9\% failure rate for at least one camera.
    \item T3: Generalization to boxes with different appearances, such as different colors and patterns (5 rollouts).
    \item T4: Introducing random camera latency, with a 10\% chance per time step and delays sampled from a uniform distribution $\mathcal{U}(0,0.5s)$ (2 rollouts).  
    \item T5: Generalization to robot appearance changes by adding googly-eye stickers all over the robot (1 rollout).
    
\end{itemize}

\vspace{2mm}\noindent\textbf{Comparisons:} We compare \ours against the following approaches in this task:
\begin{itemize}[leftmargin=4mm]
    \item \texttt{Top-down Camera}: Observations from a single, static top-down camera (Fig.~\ref{fig:baseline}).
    \item \texttt{Head Camera}: Observations from a single camera mounted at the center of the head (Fig.~\ref{fig:baseline}).
    \item \texttt{w/o Camera Pose}: A whole-body visuomotor policy trained without view-dependent positional encoding.
    \item \texttt{w/o Blink Training}: A whole-body visuomotor policy trained without randomized camera dropouts.
\end{itemize}

\vspace{2mm}\noindent\textbf{Performance:}
The training dataset contains 147 demonstration episodes, with each demonstration averaging \SI{15}{s}.
We report both qualitative and quantitative results in Fig.~\ref{fig:eval_unbox}, and illustrate typical failure cases.
\ours achieves an overall 94.4\% success rate, outperforming all baselines.

The \texttt{Top-down Camera} policy typically fails to locate the hole due to occlusion. While the robot correctly moves towards the box, it often approaches the hole too high or too low. Since the hole cannot be seen from the top-down camera, it is ambiguous at which height the robot should reach forward.

The \texttt{Head Camera} policy fails consistently to move towards the box. It always executes a fixed trajectory with the robot moving forward, independent of where the box is located. We hypothesize that this is because the box often cannot be seen from the head camera and, thus, the policy has overfitted to use proprioception instead. During our experiments, the policy only succeeded once, entering the box by coincidence and managing to open the lid smoothly since this action mostly relies on proprioception.

\texttt{w/o Camera Pose}, the policy is less accurate in locating and getting inside the hole. This demonstrates that our view-dependent positional encoding design enables better spatial awareness and more effective skill learning. And even if the robot has already gotten inside, it often fails to open the lid. We suspect that this is because vision tokens outnumber proprioception tokens, making it more challenging for the policy to learn how to utilize proprioception information.

\texttt{w/o Blink Training} policy performs worse in T2 when we randomly drop out cameras and is slightly less robust than ours in other scenarios as well. This proves that our blink training strategy is critical to the robustness of the policy, especially during unexpected test-time sensor failures.

\subsection{Sweeping Task}
\vspace{2mm}\noindent\textbf{Task:}
The robot needs to sweep all objects (small or large, randomly placed on a table or under a shelf) into a target region around its base. The target zone is $20\times20$\SI{}{cm} and centered around the robot. Training data includes demonstrations with 24 small blocks and two cylindrical objects. The task success rate for sweeping multiple small objects is measured by the ratio of objects inside the target zone. For sweeping a large object, the task success rate is measured by whether the object is dragged into the target zone without being knocked down.


\vspace{2mm}\noindent\textbf{Challenges:}
\ul{Multiple contacts:} Efficiently sweeping multiple objects requires the robot to leverage its whole body and additional inter-object contact. The robot can also sweep objects larger than itself by wrapping around them and dragging them with multiple body links.
\ul{Dexterity requirements:} The robot must conform to objects, sweep under narrow spaces (\ie under the shelf), and lift itself to varying heights to sweep tall objects while preventing them from toppling over.
\ul{Visual occlusions:} Objects under the shelf or scattered around the robot are frequently occluded from certain viewpoints.

\vspace{1mm}\noindent\textbf{Test scenarios:} We evaluate this task across 18 rollouts, divided into the following scenarios:
\begin{itemize}[leftmargin=4mm]
    \item T1: 24 small blocks, seen during training (4 rollouts).
    \item T2: Unseen small objects (4 rollouts).
    \item T3: A single, large training object (4 rollouts).
    \item T4: A single, large but unseen object (4 rollouts).
    \item T5: A mixture of small and large objects (2 rollouts).
\end{itemize}

\vspace{1mm}\noindent\textbf{Comparisons:} For this task, we compare \ours to the following approaches: 
\begin{itemize}[leftmargin=4mm]
    \item \texttt{Top-down Camera}: Observations from a single, static top-down camera (Fig.~\ref{fig:baseline}~a).
    \item \texttt{Neck Cameras}: Observations from four cameras, mounted along the robot's neck (Fig.~\ref{fig:baseline}~b).
    \item \texttt{ResNet Encoder}: A whole-body visuomotor policy with a ResNet-34~\cite{he2016deep} vision encoder trained from scratch instead of using a pretrained vision encoder.
    \item \texttt{w/o Camera Pose}: A whole-body visuomotor policy trained without view-dependent positional encoding.
    \item \texttt{w/o Blink Training}: A whole-body visuomotor policy trained without randomized camera dropouts.
    \item \texttt{Teleop EE}: Teleoperating the robot to move small objects one by one, using only the head link. This approach is used to demonstrate the inefficiency of the end-effector-only manipulation strategy, even under a near-perfect policy (\ie, teleoperation). 
\end{itemize}

\vspace{2mm}\noindent\textbf{Performance:} 
The training dataset contains 153 demonstrations, with each demonstration averaging \SI{23}{s}.
We compare the methods' success rate and time per block in Fig.~\ref{fig:eval_sweep}~(bottom right). \ours achieves a 96.6\% success rate, outperforming all baselines.

The \texttt{Top-down Camera} policy typically misses objects under the shelf due to occlusion; especially failing to fetch objects that were not successfully swept out on the first try. It also sweeps tall objects at a low height, toppling them over. As in the unboxing task, we hypothesize this behavior is due to the ambiguous height when observed from the top-down view.

The \texttt{Neck Cameras} policy misses out-of-view objects (\eg, when objects are hidden behind the robot itself) and fails to recover in cases when, \eg, the robot pushed objects too far away. Occasionally, we observe the robot getting stuck in a pose without interacting with any object. We suspect this being due to the policy overfitting to the proprioception signal due to occlusion.

The \texttt{ResNet Encoder} policy often sweeps toward incorrect or empty regions, which we hypothesize is due to inaccurate semantic understanding in the visual representation.

The \texttt{w/o Camera Pose} policy sometimes misses a few blocks and knocks down tall objects due to inaccurate motions, verifying that our view-dependent positional encoding enhances spatial reasoning capabilities.

The \texttt{w/o Blink Training} policy is less robust than ours, \eg, the robot fails to retrieve blocks being pushed further away during interactions.

\vspace{2mm}\noindent\textbf{Efficiency:} We record the time to sweep 24 blocks from (approximately) the same initial configuration to the target zone, and report the average per-block time. With whole-body sweeping, the robot manipulates object piles much more efficiently, leveraging multiple contacts (\SI{2}{s}/block for teleoperation and \SI{3.2}{s}/block for our rollouts), as compared to using only the end-effector to move objects one by one (\SI{9.3}{s}/block for teleoperation).

\begin{figure*}[t]
    \centering
    \includegraphics[width=\linewidth]{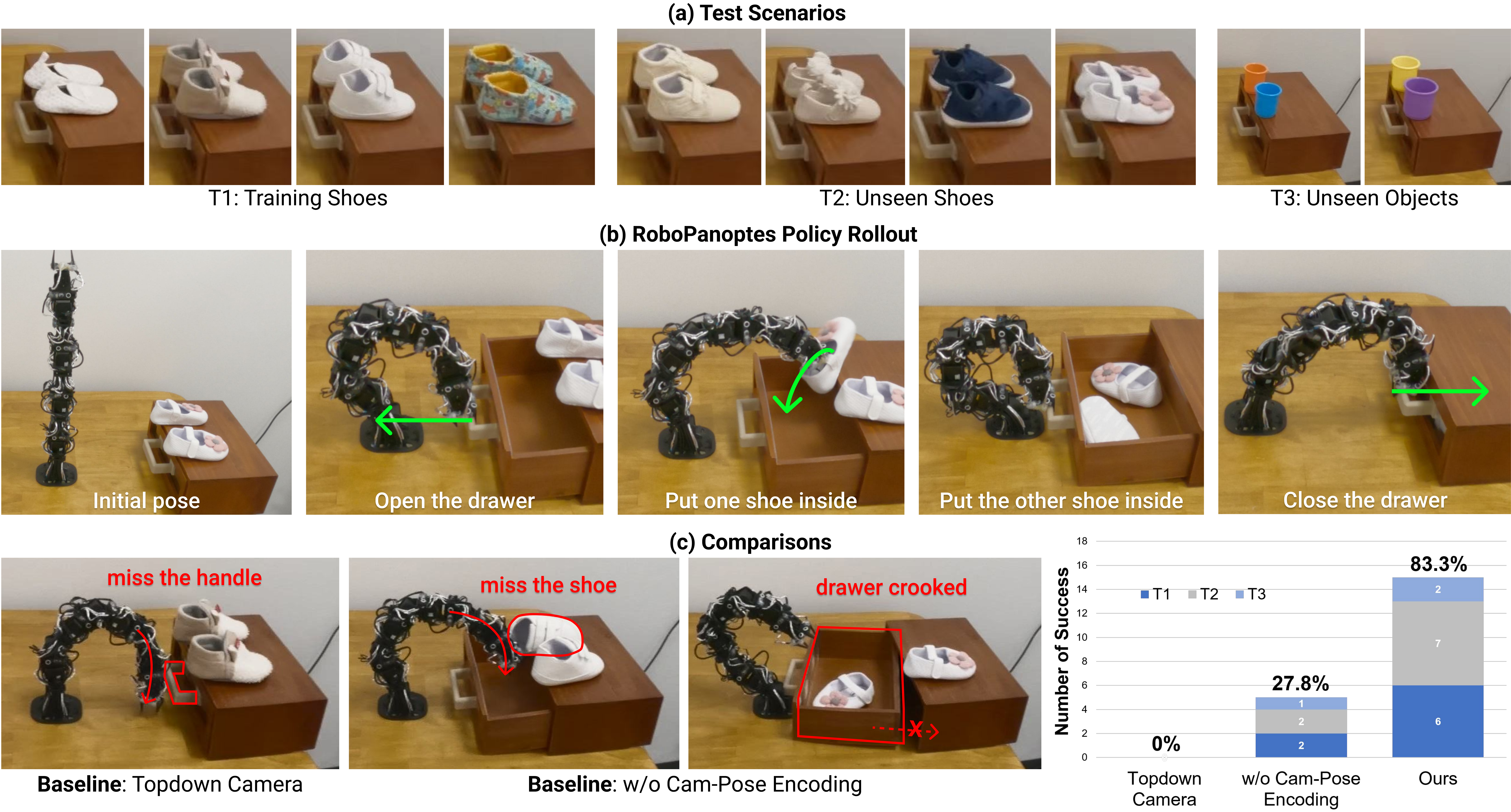}
    \caption{
    \textbf{Stowing Task.} a) Different test scenarios. b) \ours policy rollouts, demonstrating the ability of precise long-horizon manipulation in cluttered environments. c) Typical failure cases of baselines. The \texttt{Top-down Camera} policy fails to locate the handle. The \texttt{w/o Camera Pose Encoding} policy's actions are less precise, leading to failures like missing the shoe or misaligning the drawer.}
    \label{fig:eval_stow}
    \vspace{-4mm}
\end{figure*}

\subsection{Stowing Task}
\vspace{2mm}\noindent\textbf{Task:}
As shown in Fig.~\ref{fig:eval_stow}, in this task, the robot performs a sequence of actions to stow shoes in a drawer. Specifically, it hooks and pulls the drawer handle to open it, picks up a pair of shoes (one by one), places the shoes inside the drawer, and pushes the drawer to close it. The take success rate is measured by whether both shoes are inside the drawer and the drawer is fully closed in the end.

\vspace{2mm}\noindent\textbf{Challenges:}
\ul{Long horizon:} This task requires long-horizon sequential actions, involving multiple articulated and deformable objects.
\ul{Precision:} Hooking the handle and placing shoes inside the drawer requires exact manipulation.
\ul{Cluttered environment:} The robot must navigate constrained drawer space and avoid collisions. Also, the robot picks up the cluttered shoes one by one.

\vspace{2mm}\noindent\textbf{Test scenarios:} We evaluate the task over 18 rollouts, divided as follows (Fig.~\ref{fig:eval_stow} (a)):
\begin{itemize}[leftmargin=4mm]
    \item T1: Stowing 4 pairs of shoes, seen during training, and each pair in two different initial configurations (8 rollouts).
    \item T2: Stowing 4 pairs of unseen shoes (8 rollouts).
    \item T3: Stowing unseen objects (\ie, cups) (2 rollouts).
\end{itemize}
We compare the \ours with two systems described before, one using only a top-down camera and one without view-dependent positional encoding. 

\vspace{2mm}\noindent\textbf{Performance:}
The training dataset contains 90 demonstrations, with each demonstration averaging \SI{20}{s}.
We report the qualitative and quantitative results in Fig.~\ref{fig:eval_stow}, and illustrate typical failure cases.
\ours achieves an overall success rate of 83.3\%, compared to 27.8\% for the \texttt{w/o Camera Pose} policy and 0\% for the \texttt{Top-down Camera} policy (Fig.~\ref{fig:eval_stow}). 
Since our demonstration data contains behaviors of recovering from a sub-goal failure (\eg, failed grasps), we observe that the learned policy is able to capture this kind of robust retry-and-recovery behaviors. 
These behaviors are best viewed in our supplementary videos.


The \texttt{Top-down Camera} policy fails at the first step of opening the drawer, missing the handle. We conjecture that this is due to the handle being occluded when the robot is bending forward, and the ambiguity of the required height when viewed from top down.

The \texttt{w/o Camera Pose} policy's actions are less precise, leading to behaviors such as missing the shoes and pulling the drawer out too much. This might be due to the complexity of learning from blurry observations, resulting from the constantly moving cameras. Supporting this hypothesis, we observe that the policy especially struggles when the robot is going through large whole-body motions, such as bending a large degree to open or close the drawer.









\section{Limitations and future work}
\label{sec:limitation}
Our system experiences three most common failure cases: 1) In the unboxing task, certain DYNAMIXEL motors automatically shut down after rapid movement due to overheating, preventing complete lid opening; 2) In sweeping, blocks occasionally move beyond the robot's reach range; 3) When stowing diverse-shaped shoes, motion imprecision can cause misalignment in cluttered spaces, preventing drawer closure. Using stronger and more precise motors could improve system performance.

In our current implementation, the robot's fixed base limits it to table-top manipulation tasks.
A future direction is to extend the current setup to a mobile manipulation system, increasing the robot's operation range while enabling capabilities in challenging in-the-wild scenarios. This would require modifications to the data collection system to enable the demonstration of combined locomotion and manipulation, as well as removing the wires that tether the robot to the PC by moving to on-board computation. 
Future mobile systems could potentially use small-sized camera development boards with WIFI/Bluetooth that can work wirelessly, \eg, \texttt{ESP32CAM}.
%
Another future direction would be to automatically optimize morphology design and sensor placement with generative models~\cite{xu2024dynamics}.

\section{Conclusion} 
\label{sec:conclusion}

\ours demonstrates the effectiveness of whole-body vision and dexterity through an integrated robot system. Our system goes beyond traditional robotic manipulation by leveraging all of its body parts to sense and interact with its surroundings.   
Using a whole-body visuomotor policy, \ours learns to infer complex whole-body actions from high-dimensional camera observations, while remaining robust to potential sensor failures. 
Extensive real-world experiments suggest that our proposed system increases robotic manipulation capability and efficiency on a wide range of challenging tasks. 
We hope that the presented hardware design, learning framework, and experimental results will encourage future exploration of unconventional manipulation systems such as whole-body vision and dexterity. 
%

%



\section*{Acknowledgement}

The authors would like to thank Han Zhang, Ken Wang, Yifan Hou, Haoyu Xiong, Haochen Shi, Mengda Xu, Cheng Chi, Zhenjia Xu, and Adrian Wong for their advice on the design of the robot and implementation of hardware. We thank Huy Ha for his help with visualizations and video editing. In addition, we would like to thank all REALab members: Mandi Zhao, Zeyi Liu, Max Du, Hojung Choi, Austin Patel, Shuang Li, Chuer Pan, Yihuai Gao, John So, and Eric Liang for fruitful discussions.
This work was supported in part by NSF Award \#2143601, \#2037101, and \#2132519.  Dominik Bauer is partially supported by the Austrian Science Fund (FWF) under project \# J 4683. 
The views and conclusions contained herein are those of the authors and should not be
interpreted as necessarily representing the official policies, either expressed or implied, of the sponsors.

\bibliographystyle{plainnat}
\bibliography{references}

\begin{thebibliography}{48}
\providecommand{\natexlab}[1]{#1}
\providecommand{\url}[1]{\texttt{#1}}
\expandafter\ifx\csname urlstyle\endcsname\relax
  \providecommand{\doi}[1]{doi: #1}\else
  \providecommand{\doi}{doi: \begingroup \urlstyle{rm}\Url}\fi

\bibitem[Blumenschein et~al.(2020)Blumenschein, Coad, Haggerty, Okamura, and Hawkes]{blumenschein2020design}
Laura~H Blumenschein, Margaret~M Coad, David~A Haggerty, Allison~M Okamura, and Elliot~W Hawkes.
\newblock \href{https://www.frontiersin.org/articles/10.3389/frobt.2020.548266/pdf}{Design, modeling, control, and application of everting vine robots}.
\newblock \emph{Frontiers in Robotics and AI}, 7:\penalty0 548266, 2020.

\bibitem[Cai et~al.(2025)Cai, Xu, Wang, Chen, Matusik, Shou, and Chen]{cai2025modular}
Yilin Cai, Haokai Xu, Yifan Wang, Desai Chen, Wojciech Matusik, Wan Shou, and Yue Chen.
\newblock \href{https://ieeexplore.ieee.org/abstract/document/10829627}{Modular Self-Reconfigurable Continuum Robot for General Purpose Loco-Manipulation}.
\newblock \emph{IEEE Robotics and Automation Letters}, 2025.

\bibitem[Caron et~al.(2021)Caron, Touvron, Misra, J{\'e}gou, Mairal, Bojanowski, and Joulin]{caron2021emerging}
Mathilde Caron, Hugo Touvron, Ishan Misra, Herv{\'e} J{\'e}gou, Julien Mairal, Piotr Bojanowski, and Armand Joulin.
\newblock \href{https://openaccess.thecvf.com/content/ICCV2021/papers/Caron_Emerging_Properties_in_Self-Supervised_Vision_Transformers_ICCV_2021_paper.pdf}{Emerging properties in self-supervised vision transformers}.
\newblock In \emph{Proceedings of the IEEE/CVF international conference on computer vision}, pages 9650--9660, 2021.

\bibitem[Chi et~al.(2023)Chi, Xu, Feng, Cousineau, Du, Burchfiel, Tedrake, and Song]{chi2023diffusion}
Cheng Chi, Zhenjia Xu, Siyuan Feng, Eric Cousineau, Yilun Du, Benjamin Burchfiel, Russ Tedrake, and Shuran Song.
\newblock \href{https://journals.sagepub.com/doi/pdf/10.1177/02783649241273668}{Diffusion policy: Visuomotor policy learning via action diffusion}.
\newblock \emph{The International Journal of Robotics Research}, page 02783649241273668, 2023.

\bibitem[Chi et~al.(2024)Chi, Xu, Pan, Cousineau, Burchfiel, Feng, Tedrake, and Song]{chi2024universal}
Cheng Chi, Zhenjia Xu, Chuer Pan, Eric Cousineau, Benjamin Burchfiel, Siyuan Feng, Russ Tedrake, and Shuran Song.
\newblock \href{https://arxiv.org/pdf/2402.10329}{Universal manipulation interface: In-the-wild robot teaching without in-the-wild robots}.
\newblock \emph{arXiv preprint arXiv:2402.10329}, 2024.

\bibitem[Coad et~al.(2019)Coad, Blumenschein, Cutler, Zepeda, Naclerio, El-Hussieny, Mehmood, Ryu, Hawkes, and Okamura]{coad2019vine}
Margaret~M Coad, Laura~H Blumenschein, Sadie Cutler, Javier A~Reyna Zepeda, Nicholas~D Naclerio, Haitham El-Hussieny, Usman Mehmood, Jee-Hwan Ryu, Elliot~W Hawkes, and Allison~M Okamura.
\newblock \href{https://ieeexplore.ieee.org/iel7/100/9194250/08917931.pdf}{Vine robots}.
\newblock \emph{IEEE Robotics \& Automation Magazine}, 27\penalty0 (3):\penalty0 120--132, 2019.

\bibitem[Dahiya et~al.(2009)Dahiya, Metta, Valle, and Sandini]{dahiya2009tactile}
Ravinder~S Dahiya, Giorgio Metta, Maurizio Valle, and Giulio Sandini.
\newblock \href{https://ieeexplore.ieee.org/iel5/8860/5406215/05339133.pdf}{Tactile sensing—from humans to humanoids}.
\newblock \emph{IEEE transactions on robotics}, 26\penalty0 (1):\penalty0 1--20, 2009.

\bibitem[Dean-Leon et~al.(2019)Dean-Leon, Guadarrama-Olvera, Bergner, and Cheng]{dean2019whole}
Emmanuel Dean-Leon, J~Rogelio Guadarrama-Olvera, Florian Bergner, and Gordon Cheng.
\newblock \href{https://ieeexplore.ieee.org/iel7/8780387/8793254/08793258.pdf}{Whole-body active compliance control for humanoid robots with robot skin}.
\newblock In \emph{2019 International Conference on Robotics and Automation (ICRA)}, pages 5404--5410. IEEE, 2019.

\bibitem[Dosovitskiy(2020)]{dosovitskiy2020image}
Alexey Dosovitskiy.
\newblock \href{https://arxiv.org/pdf/2010.11929}{An image is worth 16x16 words: Transformers for image recognition at scale}.
\newblock \emph{arXiv preprint arXiv:2010.11929}, 2020.

\bibitem[Elsayed et~al.(2021)Elsayed, Takemori, Tanaka, and Matsuno]{elsayed2021mobile}
Belal~A Elsayed, Tatsuya Takemori, Motoyasu Tanaka, and Fumitoshi Matsuno.
\newblock \href{https://ieeexplore.ieee.org/iel7/3516/9924618/09562974.pdf}{Mobile manipulation using a snake robot in a helical gait}.
\newblock \emph{IEEE/ASME Transactions on Mechatronics}, 27\penalty0 (5):\penalty0 2600--2611, 2021.

\bibitem[Fu et~al.(2023)Fu, Cheng, and Pathak]{fu2023deep}
Zipeng Fu, Xuxin Cheng, and Deepak Pathak.
\newblock \href{https://proceedings.mlr.press/v205/fu23a/fu23a.pdf}{Deep whole-body control: learning a unified policy for manipulation and locomotion}.
\newblock In \emph{Conference on Robot Learning}, pages 138--149. PMLR, 2023.

\bibitem[Fu et~al.(2024)Fu, Zhao, and Finn]{fu2024mobile}
Zipeng Fu, Tony~Z Zhao, and Chelsea Finn.
\newblock \href{https://arxiv.org/pdf/2401.02117.pdf?trk=public_post_comment-text}{Mobile aloha: Learning bimanual mobile manipulation with low-cost whole-body teleoperation}.
\newblock \emph{arXiv preprint arXiv:2401.02117}, 2024.

\bibitem[Goncalves et~al.(2022)Goncalves, Kuppuswamy, Beaulieu, Uttamchandani, Tsui, and Alspach]{goncalves2022punyo}
Aimee Goncalves, Naveen Kuppuswamy, Andrew Beaulieu, Avinash Uttamchandani, Katherine~M Tsui, and Alex Alspach.
\newblock \href{https://ieeexplore.ieee.org/iel7/9762008/9762065/09762117.pdf}{Punyo-1: Soft tactile-sensing upper-body robot for large object manipulation and physical human interaction}.
\newblock In \emph{2022 IEEE 5th International Conference on Soft Robotics (RoboSoft)}, pages 844--851. IEEE, 2022.

\bibitem[Ha et~al.(2024)Ha, Gao, Fu, Tan, and Song]{ha2024umi}
Huy Ha, Yihuai Gao, Zipeng Fu, Jie Tan, and Shuran Song.
\newblock \href{https://arxiv.org/pdf/2407.10353}{Umi on legs: Making manipulation policies mobile with manipulation-centric whole-body controllers}.
\newblock \emph{arXiv preprint arXiv:2407.10353}, 2024.

\bibitem[He et~al.(2016)He, Zhang, Ren, and Sun]{he2016deep}
Kaiming He, Xiangyu Zhang, Shaoqing Ren, and Jian Sun.
\newblock \href{https://openaccess.thecvf.com/content_cvpr_2016/html/He_Deep_Residual_Learning_CVPR_2016_paper.html}{Deep residual learning for image recognition}.
\newblock In \emph{Proceedings of the IEEE conference on computer vision and pattern recognition}, pages 770--778, 2016.

\bibitem[Jeon et~al.(2023)Jeon, Jung, Choi, Kim, and Hwangbo]{jeon2023learning}
Seunghun Jeon, Moonkyu Jung, Suyoung Choi, Beomjoon Kim, and Jemin Hwangbo.
\newblock \href{https://ieeexplore.ieee.org/iel7/7083369/7339444/10325606.pdf}{Learning whole-body manipulation for quadrupedal robot}.
\newblock \emph{IEEE Robotics and Automation Letters}, 9\penalty0 (1):\penalty0 699--706, 2023.

\bibitem[Jiang and Wong(2024)]{jiang2024hierarchical}
Shuo Jiang and Lawson~LS Wong.
\newblock \href{https://ieeexplore.ieee.org/iel8/10609961/10609862/10610834.pdf}{A Hierarchical Framework for Robot Safety using Whole-body Tactile Sensors}.
\newblock In \emph{2024 IEEE International Conference on Robotics and Automation (ICRA)}, pages 8021--8028. IEEE, 2024.

\bibitem[Jiang et~al.(2024)Jiang, Salagame, Ramezani, and Wong]{jiang2024snake}
Shuo Jiang, Adarsh Salagame, Alireza Ramezani, and Lawson~LS Wong.
\newblock \href{https://ieeexplore.ieee.org/iel8/10609961/10609862/10611384.pdf}{Snake Robot with Tactile Perception Navigates on Large-scale Challenging Terrain}.
\newblock In \emph{2024 IEEE International Conference on Robotics and Automation (ICRA)}, pages 5090--5096. IEEE, 2024.

\bibitem[Kim et~al.(2024)Kim, Srouji, Chen, and Zhang]{kim2024armor}
Daehwa Kim, Mario Srouji, Chen Chen, and Jian Zhang.
\newblock \href{https://arxiv.org/pdf/2412.00396}{ARMOR: Egocentric Perception for Humanoid Robot Collision Avoidance and Motion Planning}.
\newblock \emph{arXiv preprint arXiv:2412.00396}, 2024.

\bibitem[Kollmitz et~al.(2018)Kollmitz, B{\"u}scher, Schubert, and Burgard]{kollmitz2018whole}
Marina Kollmitz, Daniel B{\"u}scher, Tobias Schubert, and Wolfram Burgard.
\newblock \href{https://ieeexplore.ieee.org/iel7/8449910/8460178/08460510.pdf}{Whole-body sensory concept for compliant mobile robots}.
\newblock In \emph{2018 IEEE International Conference on Robotics and Automation (ICRA)}, pages 5429--5435. IEEE, 2018.

\bibitem[Liljeb{\"a}ck et~al.(2012)Liljeb{\"a}ck, Pettersen, Stavdahl, and Gravdahl]{liljeback2012review}
P{\aa}l Liljeb{\"a}ck, Kristin~Ytterstad Pettersen, {\O}yvind Stavdahl, and Jan~Tommy Gravdahl.
\newblock \href{https://www.sciencedirect.com/science/article/pii/S0921889011001618}{A review on modelling, implementation, and control of snake robots}.
\newblock \emph{Robotics and Autonomous systems}, 60\penalty0 (1):\penalty0 29--40, 2012.

\bibitem[Liu et~al.(2021{\natexlab{a}})Liu, Tong, and Liu]{liu2021review}
Jindong Liu, Yuchuang Tong, and Jinguo Liu.
\newblock \href{https://www.sciencedirect.com/science/article/pii/S0921889021000701}{Review of snake robots in constrained environments}.
\newblock \emph{Robotics and Autonomous Systems}, 141:\penalty0 103785, 2021{\natexlab{a}}.

\bibitem[Liu et~al.(2021{\natexlab{b}})Liu, Balatti, Ellis, Hadjivelichkov, Stoyanov, Ajoudani, and Kanoulas]{liu2021garbage}
Jingyi Liu, Pietro Balatti, Kirsty Ellis, Denis Hadjivelichkov, Danail Stoyanov, Arash Ajoudani, and Dimitrios Kanoulas.
\newblock \href{https://ieeexplore.ieee.org/iel7/9555758/9555669/09555800.pdf}{Garbage collection and sorting with a mobile manipulator using deep learning and whole-body control}.
\newblock In \emph{2020 IEEE-RAS 20th International Conference on Humanoid Robots (Humanoids)}, pages 408--414. IEEE, 2021{\natexlab{b}}.

\bibitem[Liu et~al.(2023{\natexlab{a}})Liu, Jing, Huang, Dun, Qiao, Leung, and Chen]{liu2023inchworm}
Wuji Liu, Zhongliang Jing, Jianzhe Huang, Xiangming Dun, Lingfeng Qiao, Henry Leung, and Wujun Chen.
\newblock \href{https://ieeexplore.ieee.org/iel7/41/4387790/10034485.pdf}{An Inchworm-Snake Inspired Flexible Robotic Manipulator With Multisection SMA Actuators for Object Grasping}.
\newblock \emph{IEEE Transactions on Industrial Electronics}, 70\penalty0 (12):\penalty0 12616--12625, 2023{\natexlab{a}}.

\bibitem[Liu et~al.(2023{\natexlab{b}})Liu, Xu, Chen, Yuan, Wang, Xu, Chen, and Yi]{liu2023enhancing}
Yun Liu, Xiaomeng Xu, Weihang Chen, Haocheng Yuan, He~Wang, Jing Xu, Rui Chen, and Li~Yi.
\newblock \href{https://ieeexplore.ieee.org/iel7/7083369/7339444/10333330.pdf}{Enhancing generalizable 6d pose tracking of an in-hand object with tactile sensing}.
\newblock \emph{IEEE Robotics and Automation Letters}, 2023{\natexlab{b}}.

\bibitem[Qi et~al.(2022)Qi, Song, and Dai]{qi2022safe}
Keke Qi, Zhibin Song, and Jian~S Dai.
\newblock \href{https://www.sciencedirect.com/science/article/pii/S0736584521001605}{Safe physical human-robot interaction: A quasi whole-body sensing method based on novel laser-ranging sensor ring pairs}.
\newblock \emph{Robotics and Computer-Integrated Manufacturing}, 75:\penalty0 102280, 2022.

\bibitem[Qin et~al.(2022)Qin, Wu, Cheng, Pan, Zhao, Shi, Song, and Ji]{qin2022adaptive}
Guodong Qin, Huapeng Wu, Yong Cheng, Hongtao Pan, Wenlong Zhao, Shanshuang Shi, Yuntao Song, and Aihong Ji.
\newblock \href{https://www.sciencedirect.com/science/article/pii/S0307904X22000646}{Adaptive trajectory control of an under-actuated snake robot}.
\newblock \emph{Applied Mathematical Modelling}, 106:\penalty0 756--769, 2022.

\bibitem[Radford et~al.(2021)Radford, Kim, Hallacy, Ramesh, Goh, Agarwal, Sastry, Askell, Mishkin, Clark, et~al.]{radford2021learning}
Alec Radford, Jong~Wook Kim, Chris Hallacy, Aditya Ramesh, Gabriel Goh, Sandhini Agarwal, Girish Sastry, Amanda Askell, Pamela Mishkin, Jack Clark, et~al.
\newblock \href{http://proceedings.mlr.press/v139/radford21a/radford21a.pdf}{Learning transferable visual models from natural language supervision}.
\newblock In \emph{International conference on machine learning}, pages 8748--8763. PMLR, 2021.

\bibitem[Ren et~al.(2022)Ren, Liu, Hu, and Li]{ren2022integrated}
Xiaoqian Ren, Yueyue Liu, Yingbai Hu, and Zhijun Li.
\newblock \href{https://ieeexplore.ieee.org/iel7/8856/4358066/09732182.pdf}{Integrated task sensing and whole body control for mobile manipulation with series elastic actuators}.
\newblock \emph{IEEE Transactions on Automation Science and Engineering}, 20\penalty0 (1):\penalty0 413--424, 2022.

\bibitem[Salagame et~al.(2024)Salagame, Gangaraju, Nallaguntla, Sihite, Schirner, and Ramezani]{salagame2024loco}
Adarsh Salagame, Kruthika Gangaraju, Harin~Kumar Nallaguntla, Eric Sihite, Gunar Schirner, and Alireza Ramezani.
\newblock \href{https://arxiv.org/pdf/2404.08174}{Loco-Manipulation with Nonimpulsive Contact-Implicit Planning in a Slithering Robot}.
\newblock \emph{arXiv preprint arXiv:2404.08174}, 2024.

\bibitem[Seeja et~al.(2022)Seeja, Doss, and Hency]{seeja2022survey}
G~Seeja, Arockia Selvakumar~Arockia Doss, and V~Berlin Hency.
\newblock \href{https://ieeexplore.ieee.org/iel7/6287639/6514899/09921291.pdf}{A survey on snake robot locomotion}.
\newblock \emph{IEEE Access}, 10:\penalty0 112100--112116, 2022.

\bibitem[Sferrazza et~al.(2024)Sferrazza, Huang, Lin, Lee, and Abbeel]{sferrazza2024humanoidbench}
Carmelo Sferrazza, Dun-Ming Huang, Xingyu Lin, Youngwoon Lee, and Pieter Abbeel.
\newblock \href{https://arxiv.org/pdf/2403.10506}{Humanoidbench: Simulated humanoid benchmark for whole-body locomotion and manipulation}.
\newblock \emph{arXiv preprint arXiv:2403.10506}, 2024.

\bibitem[Shaw et~al.(2024)Shaw, Li, Yang, Srirama, Liu, Xiong, Mendonca, and Pathak]{shaw2024bimanual}
Kenneth Shaw, Yulong Li, Jiahui Yang, Mohan~Kumar Srirama, Ray Liu, Haoyu Xiong, Russell Mendonca, and Deepak Pathak.
\newblock \href{https://arxiv.org/pdf/2411.13677}{Bimanual Dexterity for Complex Tasks}.
\newblock In \emph{8th Annual Conference on Robot Learning}, 2024.

\bibitem[Skand et~al.()Skand, Pandit, Kim, Fuxin, and Lee]{skandsimple}
Skand Skand, Bikram Pandit, Chanho Kim, Li~Fuxin, and Stefan Lee.
\newblock \href{https://openreview.net/pdf?id=AsbyZRdqPv}{Simple Masked Training Strategies Yield Control Policies That Are Robust to Sensor Failure}.
\newblock In \emph{8th Annual Conference on Robot Learning}.

\bibitem[Stasse and Righetti(2020)]{stasse:hal-03045448}
Olivier Stasse and Ludovic Righetti.
\newblock \href{https://hal.science/hal-03045448}{Whole-Body Manipulation}.
\newblock In \emph{{Encyclopedia of Robotics}}, pages 1--9. {Springer Berlin Heidelberg}, December 2020.
\newblock \doi{10.1007/978-3-642-41610-1\_187-1}.

\bibitem[Tanaka et~al.(2015)Tanaka, Kon, and Tanaka]{tanaka2015range}
Motoyasu Tanaka, Kazuyuki Kon, and Kazuo Tanaka.
\newblock \href{https://ieeexplore.ieee.org/iel7/87/7185486/07027788.pdf}{Range-sensor-based semiautonomous whole-body collision avoidance of a snake robot}.
\newblock \emph{IEEE Transactions on Control Systems Technology}, 23\penalty0 (5):\penalty0 1927--1934, 2015.

\bibitem[Uppal et~al.(2024)Uppal, Agarwal, Xiong, Shaw, and Pathak]{uppal2024spin}
Shagun Uppal, Ananye Agarwal, Haoyu Xiong, Kenny Shaw, and Deepak Pathak.
\newblock \href{https://openaccess.thecvf.com/content/CVPR2024/papers/Uppal_SPIN_Simultaneous_Perception_Interaction_and_Navigation_CVPR_2024_paper.pdf}{SPIN: Simultaneous Perception, Interaction and Navigation}.
\newblock \emph{CVPR}, 2024.

\bibitem[Vaswani(2017)]{vaswani2017attention}
A~Vaswani.
\newblock \href{https://arxiv.org/pdf/1706.03762}{Attention is all you need}.
\newblock \emph{Advances in Neural Information Processing Systems}, 2017.

\bibitem[Wang et~al.(2023)Wang, Pierce, Kojouharov, Chong, Diaz, Lu, and Goldman]{wang2023mechanical}
Tianyu Wang, Christopher Pierce, Velin Kojouharov, Baxi Chong, Kelimar Diaz, Hang Lu, and Daniel~I Goldman.
\newblock \href{https://www.science.org/doi/pdf/10.1126/scirobotics.adi2243}{Mechanical intelligence simplifies control in terrestrial limbless locomotion}.
\newblock \emph{Science Robotics}, 8\penalty0 (85):\penalty0 eadi2243, 2023.

\bibitem[Wright et~al.(2007)Wright, Johnson, Peck, McCord, Naaktgeboren, Gianfortoni, Gonzalez-Rivero, Hatton, and Choset]{wright2007design}
Cornell Wright, Aaron Johnson, Aaron Peck, Zachary McCord, Allison Naaktgeboren, Philip Gianfortoni, Manuel Gonzalez-Rivero, Ross Hatton, and Howie Choset.
\newblock \href{https://ieeexplore.ieee.org/iel5/4398943/4398944/04399617.pdf}{Design of a modular snake robot}.
\newblock In \emph{2007 IEEE/RSJ International Conference on Intelligent Robots and Systems}, pages 2609--2614. IEEE, 2007.

\bibitem[Wu et~al.(2024)Wu, Shentu, Yi, Lin, and Abbeel]{wu2024gello}
Philipp Wu, Yide Shentu, Zhongke Yi, Xingyu Lin, and Pieter Abbeel.
\newblock \href{https://ieeexplore.ieee.org/abstract/document/10801581}{Gello: A general, low-cost, and intuitive teleoperation framework for robot manipulators}.
\newblock In \emph{2024 IEEE/RSJ International Conference on Intelligent Robots and Systems (IROS)}, pages 12156--12163. IEEE, 2024.

\bibitem[Xiong et~al.(2024)Xiong, Mendonca, Shaw, and Pathak]{xiong2024adaptive}
Haoyu Xiong, Russell Mendonca, Kenneth Shaw, and Deepak Pathak.
\newblock \href{https://arxiv.org/pdf/2401.14403}{Adaptive Mobile Manipulation for Articulated Objects In the Open World}.
\newblock \emph{arXiv preprint arXiv:2401.14403}, 2024.

\bibitem[Xu et~al.(2024{\natexlab{a}})Xu, Xu, Xu, Chi, Wetzstein, Veloso, and Song]{xu2024flow}
Mengda Xu, Zhenjia Xu, Yinghao Xu, Cheng Chi, Gordon Wetzstein, Manuela Veloso, and Shuran Song.
\newblock \href{https://arxiv.org/pdf/2407.15208}{Flow as the cross-domain manipulation interface}.
\newblock \emph{arXiv preprint arXiv:2407.15208}, 2024{\natexlab{a}}.

\bibitem[Xu et~al.(2023)Xu, Yang, Mo, Pan, Yi, and Guibas]{xu2023jacobinerf}
Xiaomeng Xu, Yanchao Yang, Kaichun Mo, Boxiao Pan, Li~Yi, and Leonidas Guibas.
\newblock \href{http://openaccess.thecvf.com/content/CVPR2023/papers/Xu_JacobiNeRF_NeRF_Shaping_With_Mutual_Information_Gradients_CVPR_2023_paper.pdf}{Jacobinerf: Nerf shaping with mutual information gradients}.
\newblock In \emph{Proceedings of the IEEE/CVF Conference on Computer Vision and Pattern Recognition}, pages 16498--16507, 2023.

\bibitem[Xu et~al.(2024{\natexlab{b}})Xu, Ha, and Song]{xu2024dynamics}
Xiaomeng Xu, Huy Ha, and Shuran Song.
\newblock Dynamics-guided diffusion model for robot manipulator design.
\newblock \emph{arXiv preprint arXiv:2402.15038}, 2024{\natexlab{b}}.

\bibitem[Yamaguchi and Atkeson(2017)]{yamaguchi2017optical}
Akihiko Yamaguchi and Christopher~G Atkeson.
\newblock \href{https://www.researchgate.net/profile/Akihiko-Yamaguchi/publication/321760369_Optical_Skin_For_Robots_Tactile_Sensing_And_Whole-Body_Vision/links/5a30ea03458515a13d857486/Optical-Skin-For-Robots-Tactile-Sensing-And-Whole-Body-Vision.pdf}{Optical skin for robots: Tactile sensing and whole-body vision}.
\newblock In \emph{Workshop on Tactile Sensing for Manipulation, Robotics: Science and Systems (RSS)}, volume~25, pages 133--134, 2017.

\bibitem[Zhang et~al.(2023)Zhang, Li, Kan, Yuan, Rajabi, Wu, Peng, and Wu]{zhang2023preprogrammable}
Jie Zhang, You Li, Ziyun Kan, Qiufeng Yuan, Hamed Rajabi, Zhigang Wu, Haijun Peng, and Jianing Wu.
\newblock \href{https://www.liebertpub.com/doi/pdf/10.1089/soro.2022.0048}{A preprogrammable continuum robot inspired by elephant trunk for dexterous manipulation}.
\newblock \emph{Soft Robotics}, 10\penalty0 (3):\penalty0 636--646, 2023.

\bibitem[Zhou et~al.(2019)Zhou, Barnes, Lu, Yang, and Li]{zhou2019continuity}
Yi~Zhou, Connelly Barnes, Jingwan Lu, Jimei Yang, and Hao Li.
\newblock \href{https://openaccess.thecvf.com/content_CVPR_2019/papers/Zhou_On_the_Continuity_of_Rotation_Representations_in_Neural_Networks_CVPR_2019_paper.pdf}{On the continuity of rotation representations in neural networks}.
\newblock In \emph{Proceedings of the IEEE/CVF conference on computer vision and pattern recognition}, pages 5745--5753, 2019.

\end{thebibliography}

\end{document}